\documentclass{clv3}

\usepackage[table]{xcolor}
\definecolor{darkblue}{rgb}{0, 0, 0.5}

\usepackage[T1]{fontenc}

\usepackage[utf8]{inputenc}

\usepackage{microtype}
\usepackage{enumitem}
\usepackage{amsthm}
\usepackage{amsmath}
\usepackage{mathtools}
\usepackage{amssymb}
\usepackage{multirow}
\usepackage{comment}
\usepackage{array}
\usepackage{tabularx}
\definecolor{darkspringgreen}{rgb}{0.09, 0.45, 0.27}
\usepackage{graphicx}
\usepackage[singlelinecheck=false,justification=justified,]{caption}
\usepackage{subcaption}
\usepackage{xcolor, colortbl}
\usepackage{setspace}
\usepackage{scalerel,stackengine}
\stackMath
\newcommand\reallywidehat[1]{%
\savestack{\tmpbox}{\stretchto{%
  \scaleto{%
    \scalerel*[\widthof{\ensuremath{#1}}]{\kern-.6pt\bigwedge\kern-.6pt}%
    {\rule[-\textheight/2]{1ex}{\textheight}}
  }{\textheight}%
}{0.5ex}}%
\stackon[1pt]{#1}{\tmpbox}%
}
\makeatletter
\renewcommand\@makefntext[1]{\leftskip=2em\hskip-2em\@makefnmark#1}
\makeatother

\usepackage{algorithm,algpseudocode}
\newlength{\bibparskip}
\let\oldthebibliography\thebibliography
\renewcommand\thebibliography[1]{%
  \oldthebibliography{#1}%
  \setlength{\parskip}{\bibitemsep}%
  \setlength{\itemsep}{\bibparskip}%
}

\usepackage{enumitem}
\setlist{nolistsep}
\theoremstyle{definition}

\usepackage{xcolor}

\newcolumntype{P}[1]{>{\centering\arraybackslash}p{#1}}

\renewcommand\appendix{%
   \setcounter{section}{0}
   \setcounter{figure}{0}
   \setcounter{table}{0}
   \setcounter{equation}{0}
   \renewcommand{\thesection}{\Alph{section}}
   \renewcommand{\theequation}{\Alph{section}.\arabic{equation}}
}

\usepackage{xspace}
\newcommand{\Multi}{\textsc{Multimodal}\xspace}
\newcommand{\Combined}{\textsc{Text+Num}\xspace}

\newcommand{\Text}{\textsc{Text}\xspace}
\newcommand{\Numeric}{\textsc{Numeric}\xspace}
\newcommand{\SMARTB}{SMART\textsubscript{BERT}\xspace}
\newcommand{\SMARTR}{SMART\textsubscript{RoBERTa}\xspace}
\newcommand{\Pruthi}{\textsc{Pruthi}\xspace}
\newcommand{\Alzantot}{\textsc{Alzantot}\xspace}
\newcommand{\Textattack}{\textit{TextAttack}\xspace}
\newcommand{\Textfooler}{\textsc{Textfooler}\xspace}
\newcommand{\Bae}{\textsc{BAE}\xspace}

\newcommand{\NNIF}{\textsc{NNIF}\xspace}

\newcommand{\MR}{\textsc{MovieReview}\xspace}

\newcommand{\dknn}{\textsc{DkNN}\xspace}

\newcommand{\featpar}[1]{\textbf{\textit{#1}}}

\newcommand{\Susconstrained}{\textsc{SusConstrained}\xspace}
\definecolor{mycolor}{RGB}{202, 42, 219}

\newcommand{\diffcolour}[1]{\textcolor{blue}{#1}}

\definecolor{midgreen}{rgb}{0.2, 0.65, 0.4}

\usepackage{booktabs}
\usepackage[justification=centering]{caption}

\usepackage{supertabular}
\usepackage{adjustbox}

\usepackage{xurl}
\usepackage{pdfpages}
\usepackage{hyperref}
\usepackage{cleveref}
\hypersetup{colorlinks=true,citecolor=darkblue, linkcolor=darkblue, urlcolor=darkblue}

\begin{document}


\runningtitle{Suspiciousness of Adversarial Texts}
\runningauthor{Shakila Tonni, Pedro Faustini, Mark Dras}

\pageonefooter{Accepted to be published in Computational Linguistics.}


\title{Graded Suspiciousness of Adversarial Texts to Humans}


\author{Shakila Mahjabin Tonni\thanks{E-mail: shakila.tonni@data61.csiro.au}}
\affil{Data61, CSIRO / School of Computing, Macquarie University}

\author{Pedro Faustini}
\affil{School of Computing, Macquarie University}
\author{Mark Dras}
\affil{School of Computing, Macquarie University}

\maketitle

\begin{abstract}
Adversarial examples pose a significant challenge to deep neural networks (DNNs) across both image and text domains, with the intent to degrade model performance through carefully altered inputs. Adversarial texts, however, are distinct from adversarial images due to their requirement for semantic similarity and the discrete nature of the textual contents. This study delves into the concept of human suspiciousness, a quality distinct from the traditional focus on imperceptibility found in image-based adversarial examples, where adversarial changes are often desired to be indistinguishable to the human eye even when placed side by side with originals. 
While this is generally not possible with text, textual adversarial content must still often remain undetected or non-suspicious to human readers.
Even when the text's purpose is to deceive NLP systems or bypass filters, the text is often expected to be natural to read.

In this research, we expand the study of human suspiciousness by analyzing how individuals perceive adversarial texts. We gather and publish a novel dataset of Likert-scale human evaluations on the suspiciousness of adversarial sentences, crafted by four widely used adversarial attack methods and assess their correlation with the human ability to detect machine-generated alterations. Additionally, we develop a regression-based model to predict levels of suspiciousness and establish a baseline for future research in reducing the suspiciousness in adversarial text generation. We also demonstrate how the regressor-generated suspicious scores can be incorporated into adversarial generation methods to produce texts that are less likely to be perceived as computer-generated. 

\end{abstract}

\section{Introduction}
\label{sec:Introduction}

Adversarial examples, which maliciously aim at degrading the  performance of deep neural networks (DNNs) by carefully crafted inputs, have shown to be effective in both image and text domains
\citep[for example]{papernot2016limitations,carlini2017towards,ma2018characterizing, moosavi2016deepfool,pruthi-etal-2019-combating,alzantot-etal-2018-generating,li-etal-2020-bert-attack,garg-ramakrishnan-2020-bae}.

To generate adversarial examples, inputs for which the model's predictions are correct are targeted and modified slightly, causing the model to alter its prediction.

In the case of an image, this can be accomplished by altering only a few pixels \citep{szegedy2013intriguing, carlini2017towards}. However, generating adversarial texts differs in a number of ways: input is discrete, there is often a requirement to be semantically similar to the original, etc.
Adversarial texts are often generated by modifying the sentences either at the character level by adding, deleting, or altering one or more characters from a sentence \citep{Li2019Textbugger, pruthi-etal-2019-combating, eger-benz-2020-hero} or at the word level by altering a few of the words with semantic preserving synonymous words or tokens \citep{jin2020bert, garg-ramakrishnan-2020-bae, alzantot-etal-2018-generating}.

In adversarial images, the added noises are mostly very subtle and imperceptible to humans, with this often being an explicit goal or criterion.  However, there are ongoing discussions about what an analogous condition is for textual adversarial examples, and how necessary it is. 
Some earlier studies \citep{li-etal-2020-bert-attack} identify imperceptibility as an explicit objective, but without providing a precise definition. This goal is straightforward for images, where the original and adversarial versions can be displayed side by side for human judgment on whether any difference exists. However, this approach is not feasible for text as the differences are typically visible and obvious.
\citet{chen-etal-2022-adversarial} argued that, based on realistic scenarios where adversarial examples might be deployed, imperceptibility is not a necessary criterion, giving as an example an attempt to bypass an offensive content filter on the web while still conveying the offensive meaning.

\begin{table*}
\resizebox{\textwidth}{!}{
\begin{tabular}{ll}
\hline
original &
graced with the kind of social texture and realism that would be foreign \\
& in american teen comedies\\
\hline
adv \#1 &
\diffcolour{gaced} with the kind of social texture and realism that would be foreign \\
& in american teen comedies\\
adv \#2 &
\diffcolour{complied} with the kind of social \diffcolour{textures} and \diffcolour{lifelike} that would be foreign\\
& in \diffcolour{usa} \diffcolour{childish} comedies\\
\hline    
\end{tabular}
}
\caption{A human-authored movie review with two adversarial variants generated by \#1 \Pruthi \cite{pruthi-etal-2019-combating} and \#2 \Textfooler \cite{jin2020bert}. 
Altered words are in blue. }
\label{table:ex}
\end{table*}

However, there are human-in-the-loop scenarios where human impressions of the text are important.  Consider reviews posted to a website that aim to bypass a filter or fool a classifier but that shouldn't be dismissed by human readers as obviously computer-generated or computer-altered.  Alternatively, phishing emails have to evade spam filters but still look plausible to humans: the standard advice in cybersecurity courses is for the recipient to look for errors and other signs of poor-quality text.\footnote{\url{https://www.cisa.gov/news-events/news/avoiding-social-engineering-and-phishing-attacks}}

These sorts of scenarios led \citet{morris-etal-2020-reevaluating}, in extending to NLP the ontology of attack spaces proposed by \citet{gilmer-etal:2020}, to propose a notion of \textit{(non-)suspicion}; this is one of four kinds of constraints (along with semantics, grammaticality and overlap) that are proposed to be applied to adversarial examples.  Non-suspicion differs somewhat from imperceptibility, in that it focuses on an awareness by a human reader that a 
sentence may have been modified.  \citet{dyrmishi-etal-2023-humans} adopt this as part of their extensive analysis of adversarial example methods in NLP:

\begin{quote}
For the many NLP systems that interact with humans, we argue that effective adversarial attacks should produce valid and natural text.
Validity refers to the property humans perceive as having the same semantic properties of interest for an adversarial text as for the original text from which it was produced. Naturalness refers to the perception that humans produce an adversarial text.  Adversarial texts that are invalid and/or unnatural can still cause failed NLP model decisions, however, their ultimate effect on humans is negligible because they would fail to convey the intended meaning (e.g. hate speech that is not perceived as hateful) or
they would be suspected to be computer-generated (e.g., a phishing email using awkward vocabulary and grammar).  
\end{quote}

However, both \citet{morris-etal-2020-reevaluating} and \citet{dyrmishi-etal-2023-humans} only consider non-suspicion in a binary manner.%
\footnote{\citet{dyrmishi-etal-2023-humans}, in the above quote, use the term `naturalness', by which they are referring to a concept that strictly subsumes suspiciousness. They note that in addition to evaluating the suspiciousness notion from \citet{morris-etal-2020-reevaluating}, they investigate three related criteria under the umbrella of naturalness. In this paper, we consider only suspiciousness.}
It is clear that there are gradations in the suspiciousness with which a human might view an adversarial example.
Consider the examples in Table~\ref{table:ex}, with an original movie review and two adversarial variants of it.  
The first adversarial variant might not trigger much suspicion, a human considering that it could have been caused by a human typo; the second is much more suspicious.  
Moreover, \citet{dyrmishi-etal-2023-humans} do compile statistics aggregating their binary human judgements and then use these statistics as evidence for some adversarial methods producing more suspicious texts than others in the aggregate.  Having these graded suspiciousness annotations for \textit{individual} items permits a much wider range of possible uses than binary ones: building scorers that can assess relative likelihoods of being computer-altered, incorporating suspiciousness as a constraint in generating adversarial examples, and so on. %
\footnote{We observe that there are similar approaches to related tasks like deepfake video and audio detection \citep{sundar-etal:2021,mai-etal:2023:PLOS,groh-etal:2024}, which have moved beyond binary judgements to graded Likert-scale judgements of individual items for various purposes: understanding levels of confidence in perception or helping direct human judges to ``appropriately allocate attention to the content or perceptual cues'' \citep{groh-etal:2024}.}

In this work, we present an investigation of human judgements of the suspiciousness of adversarial texts that goes beyond the binary of the above work to reflect gradations of suspiciousness, such as might reflect differences in individual texts like those in Table~\ref{table:ex}.
First, we collect a dataset of Likert-scale suspiciousness judgements on adversarial sentences generated by four widely used methods \citep{pruthi-etal-2019-combating,alzantot-etal-2018-generating,jin2020bert,garg-ramakrishnan-2020-bae}, and analyse some characteristics of this dataset in light of observations of human accuracy in detecting machine-generated text \citep{ippolito-etal-2020-automatic,clark-etal-2021-thats}.
Second, as a baseline for future work on this task, we build a regression-based scorer from these judgements and evaluate its reliability on held-out texts as a method for predicting the suspiciousness of adversarial texts. 
Third, as an illustration of an application of such a scorer, we carry out a small-scale experiment to assess the usefulness of such a scorer in improving suspiciousness in generating adversarial texts.

The contributions of this paper are then as follows:
\begin{itemize}

\item We make available our novel dataset of Likert-scale human judgements on the suspiciousness of human and adversarial texts, along with an analysis of this data and a discussion of how this fits with prior work. \footnote{Annotated data and code: \url{https://github.com/SJabin/HumanAdvSuspiciousness}} 

\item We define and compare a range of methods for producing regression-based scorers for predicting suspicion judgements, and determine the most effective.

\item We demonstrate that such scorers can be integrated into adversarial text generation methods to generate less suspicious-looking texts.

\end{itemize}

\section{Related Works}
\label{sec:RelatedWork}

\subsection{Crafting adversarial sentences}
\label{sec:relwork-adv-methods}

Adversarial texts are generated by modifying part of an input sentence so that the output becomes incorrect \cite{goodfellow2014explaining}.
This is an area of active interest in NLP, with several surveys providing overviews of techniques \citep[for example]{zhang-etal:2020:TIST,han-etal:2022}; we briefly discuss relevant work below.  

The most common type of model attack is against classifiers, aiming to change the predicted class \citep{zhang-etal:2020:TIST}, and the most common of these kinds of adversarial texts introduce perturbations at the character level or the word level.  These are the ones we focus on in this paper, although we briefly discuss others for completeness.
Many of these types of adversarial attacks have been facilitated by the \Textattack platform \citep{morris-etal-2020-textattack}, which frames them as specific instances of a more general combinatorial optimisation task.

\paragraph{Character-level attacks}
Character-level attacks alter the characters within a sentence \citep[for example]{pruthi-etal-2019-combating, Li2019Textbugger, deepwordbug2018gao, ebrahimi-etal-2018-hotflip} by adding, removing or exchanging characters. These techniques often result in adversarial examples that lack grammatical correctness and may easily be identified by humans, if too unnatural. 

\paragraph{Word-level attacks}
Word-level attacks, on the other hand, focus on replacing words or tokens of a sentence with a synonymous word \citep[for example]{garg-ramakrishnan-2020-bae, jin2020bert, ren-etal-2019-generating, alzantot-etal-2018-generating, ribeiro-etal-2020-beyond}. Among the word-level attacks, \citet{alzantot-etal-2018-generating, jia-etal-2019-certified} ran a generic algorithm to generate adversarial text. PWWS \citep{ren-etal-2019-generating}, Particle swarm optimization (PSO) \citep{zang-etal-2020-word} and TextFooler \citep{jin2020bert} transform texts with synonyms from lexical databases and word embeddings.

More recent word-level attacks have focused more on perturbation with improved contexts and take benefit from the transformer-based Masked Language Models (MLM). A2T \citep{yoo-qi-2021-towards-improving} employs MLM word replacement to train transformer-based models (BERT, RoBERTa). BERT-Attack \citep{li-etal-2020-bert-attack} and BAE \citep{garg-ramakrishnan-2020-bae} used both word insertion and replacement with the masked language model. CLARE \cite{li-etal-2021-contextualized} allows replacing, inserting and merging of tokens to produce variable-length adversarial texts. GLARE \citep{chi-etal-2022-glare} applies generative models (GPT-2) for text infilling. CAT-Gen \citep{wang-etal-2020-cat} pre-trains an attribute classifier to alter a token from one attribute group by a token from another attribute group. 
TextCheater \citep{textcheater2023ieee} proposes a metaheuristic query-efficient search algorithm and LimeAttack \citep{hai2024limeattack} uses a local explainable method to generate adversarial texts in the hard-label context. SSCAE \citep{asl2024semantic} imposes thresholds on Universal Sentence Encoder-based and GPT-2-based semantic and syntactic correctness to generate imperceptible adversarial examples. There are a few successful attempts to attack large language models (LLMs) as well \citep{jighao2024stealthy,wang2023llm,zimu2023llm}.

\paragraph{Phrase-level attacks}
A (much smaller) third category consists of phrase-level or paraphrastic attacks. \citet{iyyer-etal-2018-adversarial} introduced syntactically controlled paraphrase networks (SCPNs) that create paraphrased adversarial sentences using a distinct constituency-based syntax format. \citet{qi-etal-2021-mind} developed a style transfer-based adversarial attack by repeatedly transforming original inputs into various text styles to generate adversarial examples; for example, the register of the text could be changed from social media posts to biblical. The Phrase-Level Textual Adversarial Attack (PLAT) \citep{lei-etal-2022-phrase} generates adversarial samples by modifying vulnerable phrases identified as attack targets by a syntactic parser, employing a pre-trained blank-infilling model. 
\citet{zou2023universaltransferableadversarialattacks} explored Large Language Model (LLM) attacks but focused on inducing objectionable generated content from LLM APIs. Similarly, \citet{you-etal-2023-large} used GPT-3.5 API to introduce diverse style-based triggers into sentences.
While these attacks represent an intriguing category of adversarial methods, they are significantly less explored compared to character-level and word-level attacks, and in general, do not suit our scenarios of interest. (For example, a review to be posted to a reviewing website with a style changed from social media to biblical would not suit here; similarly, a focus on objectionable content does not align with our goal.)
Further, they are not available on the \Textattack platform and are generally not framed in the same combinatorial optimisation manner, so we do not explore them in this paper.

\paragraph{Attacks against other tasks}
While the majority of attacks in NLP have been against classifiers since relatively early on there have been attacks against other tasks: Machine Translation \citep{belinkov2018synthetic}, Textual Entailment \citep{kang-etal-2018-adventure}, Named Entity Recognition \cite{srinivasan-vajjala-2023-multilingual}, Grammatical Error Correction (GEC) \citep{wang-zheng-2020-improving}, and so on.  Often attacks in this space are designed for the purposes of robust training or adversarial testing \citep{huang2020testing, wang-zheng-2020-improving, xiao2023leaptesting, zohreh2023blackboxtesting}, and so do not require the combinatorial optimisation framing generally necessary for the classification task \citep{lei-etal:2019:MLSys} --- for example, \citet{wang-zheng-2020-improving} produce adversarial GEC examples just by identifying `vulnerable' tokens and replacing them with grammatical errors people may make.  While interesting, these kinds of adversarial examples are out of scope for this paper.

\subsection{Adversarial text quality}

The primary goal of an adversarial attack is to change the model's prediction. Therefore, the main metrics for assessing adversarial text effectiveness revolve around the success of the attack (attack success rate) and the rate of perturbation. However, it would obviously be possible to fool the target classifier by radically changing the text, potentially to be unrecognisable.  Therefore, adversarial text quality is also considered, in terms of properties like how similar the adversarial text is to the original, how grammatical it is, and so on.  Generation methods will also include these considerations, often in the form of constraints, in their methods for producing adversarial texts.  Both evaluation and constraints often use automatic metrics to measure these properties.

\featpar{Properties measured by automatic metrics.}
\citet{morris-etal-2020-reevaluating} identified four essential types of properties for evaluating adversarial texts: semantics, grammaticality, character-level overlap with the original texts, and non-suspicion towards human readers.  We give some examples of how these properties have been operationalised.

For semantic similarity, adversarial texts are assessed on how well they preserve the semantics of the original texts. \Textfooler \citep{jin2020bert}, \Bae \citep{garg-ramakrishnan-2020-bae}, CLARE \citep{li-etal-2021-contextualized}, HQA-Attack \citep{neurips2023liu}, and GLARE \citep{chi-etal-2022-glare} assess semantic similarity using Universal Sentence Encoder (USE)-based cosine similarity. In contrast, A2T \citep{yoo-qi-2021-towards-improving} utilises a finetuned DistilBERT model for similar assessments.

For assessing grammaticality of the adversarial texts, \citet{pruthi-etal-2019-combating} deploy a word recognition model to evaluate character-level (misspelling) attacks, while \Textfooler \citep{jin2020bert}, HotFlip \citep{ebrahimi-etal-2018-hotflip}, ensemble transfer attack \citep{kwonL22transfer}, and \Bae \citep{garg-ramakrishnan-2020-bae} (in the BAE-R setting) replace words only with those of the same part of speech. Morpheus \cite{yildiz-tantug-2019-morpheus} enforces grammaticality with an inflection-constrained word swap.

Overlap properties quantify differences between original and adversarial texts.  These can be carried out at the character level through methods such as Levenshtein edit distance \citep{gao2018ieeesp} and n-gram-based measures like BLEU \citep{wang-etal-2020-cat,yildiz-tantug-2019-morpheus}, as well as word-embedding distances \citep{jin2020bert,Wang2021IGA}.

The fourth property type of \citet{morris-etal-2020-reevaluating}, non-suspicion, has broadly speaking not yet been operationalised, and is discussed below.
\\

\featpar{Human evaluation of the adversarial texts.}
Some algorithms are only evaluated by automatic metrics, as described above.  However, others include a variety of human evaluations to verify the results of the automated evaluation.  Common assessments include the validity of the adversarial perturbation, the classification accuracy of the original task by humans \citep{jin2020bert,alzantot-etal-2018-generating,garg-ramakrishnan-2020-bae,li-etal-2020-bert-attack}; the similarity of the adversarial text to the original \citep{jin2020bert,alzantot-etal-2018-generating,li-etal-2023-adversarial,li-etal-2021-contextualized,li-etal-2020-bert-attack}; and grammatical correctness \citep{jin2020bert,li-etal-2023-adversarial,li-etal-2021-contextualized,li-etal-2020-bert-attack}. 

A few works go beyond this to also ask about the naturalness of adversarial sentences. \Bae \citep{garg-ramakrishnan-2020-bae} evaluated this by asking 3 annotators to judge whether each of the original sentences and three adversarial variants is from the original distribution, on a 1--5 Likert scale over 100 samples.
PWWS \citep{ren-etal-2019-generating} measured detectability by having human readers compare original-adversarial sentence pairs.
\\

\featpar{Human perception of suspiciousness}
The aim of \citet{morris-etal-2020-reevaluating} was to propose a unified framework for evaluating adversarial texts, motivated in part by difficulty in fairly comparing some previous works.  As noted above, one of the properties they proposed for evaluation was `non-suspicion', to be evaluated in addition to semantics, grammaticality and overlap.
For the two systems they were evaluating, \Textfooler and `\textsc{GeneticAttack}' \citep{alzantot-etal-2018-generating}, in addition to asking human judges about semantic preservation (measured by a 1--5 Likert scale) and grammaticality (measured by asking judges to spot any grammatical errors), they asked about suspiciousness as follows.

\begin{table*}
\resizebox{\textwidth}{!}{

\begin{tabular}{lcc}
\toprule
& Hum./Com. & Sus.\\
Sentences & (vote) & score\\

\midrule
orig \#1 ---
the film is hampered by its predictable plot and paper thin & 9/1 & 0.1\\
supporting characters  &&\\
adv \#1 ---
the film is \diffcolour{restrained} by its predictable \diffcolour{conspiring} and paper \diffcolour{lean}& 3/7 & 0.7\\
\diffcolour{ favours qualities} &&\\\midrule

orig \#2 ---
davis is so enamored of her own creation that she can't see how  & 9/1 & 0.1\\
insufferable the character is&&\\
adv \#2 ---
davis is \diffcolour{even} enamored of her own creation that she can't & 5/5 & 0.5\\
\diffcolour{suppose} how \diffcolour{intolerable} the character is&&\\\midrule

orig \#3 ---
flat out amusing, sometimes endearing and often fabulous, with & 8/2 &0.2\\
a solid cast, noteworthy characters, delicious dialogue and & & \\
a wide supply of effective sight gags&&\\
adv \#3 ---
flat out amusing, sometimes endearing and \diffcolour{frequently mega}, with & 4/6 &0.6\\
a \diffcolour{dependable} cast, \diffcolour{prodigious} \diffcolour{roles}, \diffcolour{luscious} dialogue and & \\
a \diffcolour{wholesale} \diffcolour{affords} of \diffcolour{actionable viewpoint} gags&&\\\midrule

orig \#4 ---
payne has created a beautiful canvas, and nicholson proves &10/0&0.0\\
once again that he's the best brush in the business&&\\
adv \#4 ---
payne has \diffcolour{fabricated} a \diffcolour{sumptuous} canvas, and nicholson \diffcolour{testifies} &4/6&0.6\\
once again that he's the best brush in the business&&\\\midrule

orig \#5 ---
brian tufano's handsome widescreen photography and paul   &9/1&0.1\\
grabowsky's excellent music turn this fairly parochial melodrama into &&\\
something really rather special&&\\
adv \#5 ---
brian tufano's handsome widescreen photography and paul &9/1&0.1\\
grabowsky's excellent music turn this fairly parochial melodrama into &&\\
something really rather \diffcolour{singular}&&\\\midrule

orig \#6 ---
a formula family tearjerker told with a heavy irish brogue & 5/5&0.5\\
accentuating, rather than muting, the plot's saccharine thrust&&\\

adv \#6 ---
a formula \diffcolour{families} tearjerker told with a \diffcolour{tremendous} irish brogue & 5/5&0.5\\
accentuating, rather than muting, the \diffcolour{plots's moralizing} thrust&&\\
\bottomrule

\end{tabular}
}

\caption{Pairs of sentences written by humans and adversarial variants of those sentences produced by \Textfooler, from the \citet{morris-etal-2020-reevaluating} dataset. Hum/Com represents votes for human vs computer-altered out of the ten binary votes per sentence. Sus. score is the proportion of votes for computer-altered out of the total vote.}
\label{tab:ex_morris}
\end{table*}

To assess suspiciousness, judges had to determine whether each sentence was real or computer-altered (i.e., a binary judgement) among an even mix of real and adversarial sentences. A sentence was deemed suspicious if the percentage of judges identifying perturbed sentences as computer-altered exceeds a predetermined threshold. 
\citet{morris-etal-2020-reevaluating} only obtained judgements for \Textfooler on their sentence-length movie review dataset, as it can be a time-consuming task; for each text, they obtained 10 judgements.
Under this setup, the judges correctly nominated 69.2\% as suspicious, against the baseline of 50\% if adversarial and real were indistinguishable.

Table~\ref{tab:ex_morris} shows a few example movie reviews and their human vs computer-altered annotations by 10 human judges. Among these examples, for the sentences \#1-\#5, the original texts typically had more judges pick them as human (although not universally so), while the \Textfooler texts varied quite widely in how many judges picked them as human-written. However, in the case of sentence \#6, both the original and the adversarial sentences were chosen as human-written sentences $50\%$ of the time. Possibly, the presence of the unusual words "tearjerker", "brogue" and "accentuating" raises the suspicion of the readers. A wide range of levels of doubt about whether texts are human-authored can thus be observed in the \citet{morris-etal-2020-reevaluating} data.

\citet{dyrmishi-etal-2023-humans} expanded this evaluation by conducting an extensive human survey of 9 word-based attacks across 3 datasets, focusing on the validity and naturalness of the adversarial examples. They evaluated suspiciousness (referring back to the work of \citet{morris-etal-2020-reevaluating} here, although defining it specifically as ``the percentage of adversarial
texts recognised as `computer altered' ''), detectability, grammaticality, and meaningfulness as components of naturalness.

Their methodology involved a three-part questionnaire where an adversarial text was defined as ``a text automatically altered by a program by replacing some words with others''. In the first section, judges evaluated individual texts on validity, suspiciousness, detectability (with options to indicate if the text seems human-written or computer-altered), grammaticality, and meaningfulness, over five original and five adversarial texts without exposure to multiple versions of the same text. In the second section, to assess detectability, judges were shown four adversarial texts and given a number $n$ indicating the count of altered words, with the task of identifying $n$ modified words in the texts. The final section presented judges with both the original and adversarial versions of the same four texts they saw in the second part to assess grammatical error introduction and meaning preservation.

In terms of suspiciousness, they found that 60.33\% of the adversarial examples were perceived as computer-altered overall, which significantly exceeded the 21.43\% suspicion rate for original texts. Notably, five out of the nine adversarial attacks produce more than 25\% examples that are likely to be mislabelled by humans, indicating these examples would fail to achieve their intended effect in NLP systems checked by humans. Additionally, humans could detect nearly half (45.28\%) of the perturbed words in the adversarial texts.

Both of these works have treated suspiciousness as binary, although it is clear that there are gradations of suspiciousness both across individual examples and in the aggregate.  We use this as our starting point.

\section{Creating the Human Suspicion Dataset}
\label{sec:sus_annotation}

In this section, then, we describe the gathering and analysis of a dataset of adversarial examples annotated with graded judgements about suspiciousness.  For these graded judgements, we use a 1--5 Likert scale, as \citet{garg-ramakrishnan-2020-bae} did to evaluate what they described as ``naturalness''.  Likert scales like this are used widely in NLP, both for human judgements related to adversarial texts ---  \citet{jin2020bert} (\Textfooler) and \citet{garg-ramakrishnan-2020-bae} (BAE) use  1--5 scales for grammaticality checks, and \citet{alzantot-etal-2018-generating} assesses similarity to original texts on a 1--4 scale --- and also for standard datasets across a variety of tasks that require graded human judgements, such as the Semantic Textual Similarity (STS) dataset of \citet{cer-etal-2017-semeval} that is part of the GLUE benchmark \citep{wang-etal-2018-glue}.

We give natural language descriptions to our 5 categories to reflect levels of certainty about whether a text is one produced by a human (with a score of 1 being most human-like) or whether altered by a computer (a score of 5 being most likely to have been altered by a computer). These are presented in Table~\ref{tab:annotation}.

\begin{table}
\centering
\resizebox{.5\columnwidth}{!}{%
\begin{tabular}{@{}cl@{}}
\toprule
 Label & Description      \\ \midrule
1 &Very   confident that written by a \\
  &human with no computer alteration\\
2 &Probably written by a human with no\\
  &computer alteration\\
3 &Uncertain whether original or altered\\ 
  &by a computer   \\
4 &Probably altered by a computer \\
5 &Very confident that it is altered by \\
  &a computer \\ \bottomrule
\end{tabular}%
}
\caption{Likert-scale suspicion labels for human annotations}

\label{tab:annotation}
\end{table}

In the subsections below, we describe the data and methods we use to construct our adversarial examples and the methods for annotating them.  We then present some descriptive statistics and analyses of the collected data.

\subsection{Data gathering setup}
\label{sec:annot_setup}

\featpar{Base dataset.}  Following \citet{morris-etal-2020-reevaluating}, we also use the Rotten Tomatoes Movie Review (\MR) sentiment analysis dataset \cite{pang-lee-2005-seeing} that is widely used in the adversarial example generation literature as well \cite{pruthi-etal-2019-combating,alzantot-etal-2018-generating,ribeiro-etal-2018-semantically,ren-etal-2019-generating,iyyer-etal-2018-adversarial,yoo-qi-2021-towards-improving,li-etal-2020-bert-attack,li-etal-2021-contextualized,jin2020bert}. 
\MR is a collection of 10,662 reviews gathered from the Rotten Tomatoes movie review database.\footnote{\url{https://datarepository.wolframcloud.com/resources/Sample-Data-Movie-Review-Sentence-Polarity}} We split the dataset into 5,662 train and 5,000 test sets. It has 21 words on average and 269 words maximum per review. In all experiments, we use 128 maximum sequence lengths for the language models on \MR. 

We produce our adversarial texts using four extensively used attacks: \Pruthi \citep{pruthi-etal-2019-combating}, \Alzantot \citep{alzantot-etal-2018-generating}, \Textfooler \citep{jin2020bert} and \Bae \citep{garg-ramakrishnan-2020-bae}, via their \textit{\Textattack} library \citep{morris-etal-2020-textattack} implementations. Details of the attack methods are as follows.

\begin{itemize}
\item \Pruthi \citep{pruthi-etal-2019-combating}. This character-level attack modifies texts by randomly switching, deleting, adding characters, or introducing keyboard mistakes. In our experiments, we allow a maximum of half the words from the original text to be perturbed. 

\item \Alzantot \citep{alzantot-etal-2018-generating}. This word-level attack permits the alteration of nearly every word in a sentence using context-preserving synonyms. The synonym search is done over a large search space including the GloVe \hbox{\citep{pennington-etal-2014-glove}} and counter-fitting word vectors \citep{mrksic-etal-2016-counter}, and the Google 1 billion words LM \citep{chelba2013one}. We follow \hbox{\citet{jia-etal-2019-certified}}'s method for accelerated synonym search using a faster Transformer-XL model and limiting semantic drift by pre-selecting word neighbors. We also set the minimum perturbation limit to one-fifth of the maximum sequence length.

\item \Textfooler \citep{jin2020bert}. This is a word-level attack that ranks the words by their impact on sentence predictions, and then sources synonyms using Counter-Fitted Word Vectors \citep{mrksic-etal-2016-counter}. The method crafts a semantically equivalent sentence, controlled by the cosine similarity of their Universal Sentence Encoder (USE) encodings \citep{cer-etal-2018-universal}, and iterates this process until it can alter the target model's prediction.

\item \Bae \citep{garg-ramakrishnan-2020-bae}.This word-level attack leverages a pre-trained BERT Masked Language Model (MLM) to create adversarial sentences by masking tokens and substituting them with contextually relevant synonyms. They argue that token-level similarity-based attacks such as \Textfooler only compare word-embedding level similarity, not the overall text semantics and that this may lead to unnatural context-drifting. \Bae enforces semantic similarity using the highest Universal Sentence Encoder (USE) score \citep{cer-etal-2018-universal} and a parts-of-speech filter for the synonym replacement. This method experiments on both token replacement (R) and insertion (I) techniques. We use \Textattack \citep{morris-etal-2020-textattack} library's BAE-R (token replacement) setting. 
    
\end{itemize}
\vskip 0.1cm

\featpar{Target model.} For generating adversarial texts, we use a pre-trained BERT-base-cased model, adding a fully connected dense layer of 768 nodes, a layer of 50\% dropout, and another dense layer of 768 nodes following \citet{tonni-dras-2023-IJCNLP-learned}.
The dataset split is 80-20 train-test. We train the model for 3 epochs with $5e^{-5}$ learning rate and AdamW optimization without freezing any layer of the backbone model. This BERT\textsubscript{BASE} model achieves $90.94\%$ test accuracy on the \MR dataset on sentiment classification task. The accuracies of the clean model and the model under attack are given in Table~\ref{tab:bert_clean_adv}. We note that in all the cases, \Textfooler degrades the classifier the most, followed by \Bae.

Table \ref{tab:bert_clean_adv} also lists the number of correctly predicted test records and the generated adversarial texts from them. The BERT classifier predicts correct labels for 4350 out of 5000 test sentences. The adversarial sentences are generated on this correctly predicted test set. Among all the attacks \Textfooler is most successful in attacking the target model by reducing BERT's accuracy to only 16\%.
\\

\begin{table}
\centering
\resizebox{.6\columnwidth}{!}{%
\begin{tabular}{@{}lcc@{}}
\toprule
           & No. of sentences & Test accuracy  \\ \midrule
Original   & 4350 & 0.909          \\\midrule
\Pruthi     & 3023 & 0.518                      \\
\Alzantot   & 2999 & 0.430                      \\
\Textfooler & 4200 & 0.160                     \\
\Bae        & 3003 & 0.313                      \\ \bottomrule
\end{tabular}%
}
\caption{Number of correctly predicted original test sentences and generated adversarial sentences by BERT\textsubscript{BASE} classifier and its accuracy under attack.}
\label{tab:bert_clean_adv}
\end{table}

\featpar{Amazon Mechanical Turk (MTurk) setup.} 
Our primary annotations are collected through Amazon Mechanical Turk (MTurk).

We selected 540 original sentences and their four adversarial versions (in total 5 sets), giving 2770 sentences altogether; these were to have one human judgement each. Further, we sampled $63$ additional sentences from the 5 sets (63$\times$5 = 315 sentences); these were to have three human judgements each, giving 945 (= $3\times315$) annotations. (We refer to this latter as the `common set', and its purpose is to enable an assessment of the level of agreement.)  Altogether we collected 3715 annotations. 
The Turkers were given instructions derived from that \citet{morris-etal-2020-reevaluating}:

\begin{quote}
All the sentences are based on texts originally written by humans. However, some of them have been altered (in terms of changes in characters, words, etc) by a computer algorithm aiming to produce texts that can fool another computer system in some way.
\end{quote}

Sentences were grouped into batches of 53 sentences --- 50 sentences for our annotation purposes and 3 sentences for attention-checking purposes --- and Turkers selected a batch.%
\footnote{There was no restriction on the number of batches each Turker could accept.}
(The attention-checking sentences are excluded in the result evaluation.)
In each batch, there was an equal mix of sentences of each of the 5 types (original sentences plus 4 adversarial types of sentences). 
The batch was created in such a way that there was no repetition of the different variants of one sentence, and the Turkers would not see the same sentence twice.
The sentences within a batch were presented one at a time.
When presented with a sentence, the Turker was asked to assign a score of 1--5 to the text based on the descriptions in Table~\ref{tab:annotation}.

The reward per sentence (HIT) was 0.13 USD (equivalent to 0.20 AUD).  We chose this cost based on some small-scale informal surveys among colleagues that estimated the time allocation for completing the task was one minute per text; this would give a payment rate that meets the Australian minimum wage. Surveyors were master-level Turkers with a HIT approval rate of 95\%. The ethics approval letter was disclosed and linked with the survey.\footnote{Ethics approval by Macquarie University (Human Ethics Committee Approval Code: 5201800393).} The MTurk survey interface is illustrated in Figure \ref{fig:mturk_UI} in the appendix.

\subsection{Analyses}
\label{sec:hum_eval_analysis}

\begin{table}
\centering
\resizebox{0.7\columnwidth}{!}{%
\begin{tabular}{@{}lrrrrrc@{}}
\toprule
Sus. score  & 1    & 2    & 3   & 4   & 5   & Mean score \\\midrule\midrule
Total     & 944 & 1062 & 254 & 521 & 234 & 2.35    \\\midrule
Original    & 234  & 227 & 39  & 83 & 20  & 2.05     
\\
Adversarial & 710  & 835 & 215 & 438 & 214 & 2.42     \\\midrule
\Pruthi      & 147  & 217  & 58  & 124 & 57  & 2.55     \\
\Alzantot    & 203  & 225  & 45  & 91 & 39  & 2.23     \\
\Textfooler  & 210  & 209  & 56  & 88 & 40  & 2.24     \\
\Bae         & 150  & 184  & 56  & 135 & 78  & 2.68     \\ \bottomrule
\end{tabular}%
}
\caption{Frequency of each Likert-scale suspicion score $[1 \ldots 5]$ for original human and adversarial sentences provided to the human annotators, along with mean suspicion scores. ``Adversarial'' represents the total of all adversarial methods.}

\label{tab:sus_counts}
\end{table}

\begin{figure}
\centering
\normalsize
\includegraphics[width=0.7\columnwidth]{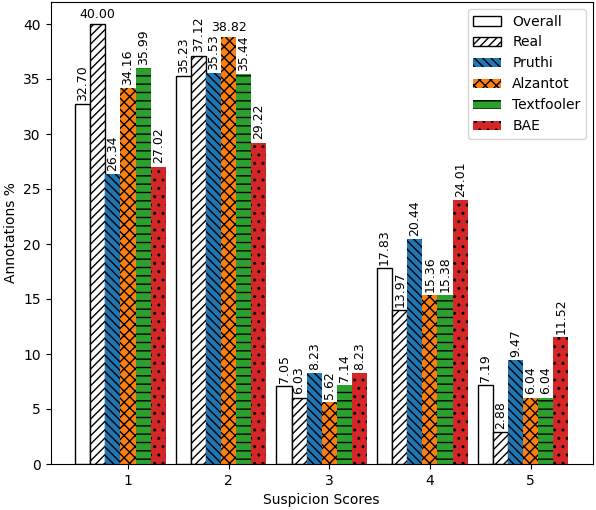}
\caption{Proportion (\%) of human suspiciousness annotations grouped by the levels of the Likert scale labels $[1 \ldots 5]$}
\label{fig:annot_dist}
\end{figure}

\featpar{Distribution of annotation scores.}
The distribution of annotation scores for all texts is given in Table~\ref{tab:sus_counts} (raw counts) and Figure~\ref{fig:annot_dist} (percentages).  Across both original texts and all adversarial texts, scores of 1 or 2 (very confident or probably written by a human) are the most common scores, with 4 (probably altered by a computer) the next most common.  The score 5 was relatively rarely chosen, indicating that annotators found indications of computer alteration to be relatively subtle. The score of 3 was likewise relatively rarely chosen, indicating that annotators did want to avoid being undecided.

This distribution of scores means that means for all text types were between 2 and 3.  As expected, original texts 
have the lowest (most human-like) scores, but the adversarial text methods all fall on the human-like side of undecided as well.  The means are significantly different using a one-way ANOVA, and a post-hoc Tukey test reveals that they fall into two groups: the human, \Alzantot and \Textfooler (with the two adversarial methods falling just outside being differentiable from human at $0.06 < p < 0.07$) as the first group; and \Pruthi and \Bae as the second group.

Following, we examine the reliability of our dataset in several ways.
\\

\featpar{Annotator agreement.} 
In light of work that shows that humans are not very good at recognising machine-generated text \citep{karpinska-etal-2021-perils,clark-etal-2021-thats}, including adversarial examples \citep{dyrmishi-etal-2023-humans}, we look to see whether there is some level of agreement about suspicion, or whether it is essentially random.  While there is a degree of common practice within the NLP community about metrics for evaluating agreement on labels representing a \textit{nominal} factor --- Cohen's kappa, Krippendorff's alpha, etc --- there is no similar degree of common practice for \textit{ordinal} factors like ours.  We, therefore, use the metric of \hbox{\citet{vogel2020likert}}, which quantifies inter-annotator agreement in Likert scale labels by calculating the average absolute deviation from the median response for each item. For $C$ annotators and $k$ Likert categories, normalised annotator disagreement of each item $i$ is:

\begin{equation}
\label{eqn:delta_i}
\delta_i = \frac{\sum_{c=1}^{C} \left| \tilde{k}_{i} - k^{c}_{i} \right|}{C}   
\end{equation}

\noindent
where $\tilde{k}_{i}$ is the median response to item $i$ and $k^{c}_{i}$ is the category supplied by annotator $c$ to item $i$.

Normalizing the $\delta_i$ by the number of items $n$ gives the average annotator disagreement level:

\begin{equation}
\label{eqn:delta}
    \delta = \frac{\sum_{i=1}^{n} \delta_i}{n}
\end{equation}

For our Likert scale of 5 points with $C=3$ annotators, the smallest possible value for $\delta_i$ is 0 (perfect agreement) and the largest possible value is 1.33.  Consider the unnormalised scores for each item $C\delta_i$: these can come from the set $\{ 0, 1, 2, 3, 4 \}$, with the largest disagreement of 4 occurring when one annotator chooses label 1 and another label 5; $\delta_i$ in this case is 1.33.  
If all items were annotated such that their $C\delta_i$ were 4, $\delta$ would likewise be 1.33.

We calculate our annotator agreement score $\delta$ over the 315 sentences receiving three judgements. Table \ref{tab:annot_disagree} presents the unnormalised per-item disagreement frequency $C\delta_i$ and the overall disagreement level $\delta$. Of the 315 sentences, unanimity ($C\delta_i=0$) was found in only 43 cases. Similar results are observed for both original and adversarial sentences. The overall $\delta$ is 0.61, a little under halfway between perfect agreement and maximal disagreement. This level of label variability aligns with humans finding this task difficult but there still is some shared intuition about the suspiciousness of texts.

The lowest level of disagreement is within the original human-authored sentences.
Within the adversarial examples, sentences altered by the  \Bae word-level attack ($\delta=0.71$) caused the most disagreement, followed by the \Pruthi character-level attack ($\delta = 0.64)$. These attacks are also the ones that have the highest mean scores in Table~\ref{tab:suspicion_level}, suggesting that these methods may generate some instances that contain perturbations that are more unusual and noticeable to some subjects, whereas the other methods are more consistent.
\\

\begin{table}
\centering
\resizebox{0.8\columnwidth}{!}{%
\begin{tabular}{@{}ccccccc@{}}
\toprule
  & Overall & Original  & \Pruthi & \Alzantot & \Textfooler & \Bae  \\ \midrule
$\delta$        & 0.61 & 0.50 & 0.64   & 0.59 & 0.61   & 0.71   \\\midrule
 $C\delta_i$=0  & 43   & 13     & 7      & 11       & 6          & 6    \\
 $C\delta_i$=1  & 104  & 24     & 20     & 24       & 26         & 10   \\
 $C\delta_i$=2  & 57   & 9      & 12     & 4        & 9          & 23   \\
 $C\delta_i$=3  & 88   & 15     & 19     & 17       & 20         & 17   \\
 $C\delta_i$=4  & 23   & 2      & 5      & 7        & 2          & 7    \\ \bottomrule
\end{tabular}%
}
\caption{ Average inter-annotator disagreements $\delta$ (Eqn~(\ref{eqn:delta})) the frequency of $C\delta_i$ (Eqn~(\ref{eqn:delta_i})) obtained from MTurk annotators. Higher scores mean more disagreement.}
\label{tab:annot_disagree}
\end{table}

\begin{table}
\centering
\begin{tabular}{@{}cccc@{}}
\toprule
      & Overall   & Original   & Adversarial   \\\midrule
$\delta$  & 0.65 & 0.62     & 0.66  \\\midrule
$C\delta_i$ = 0  & 4        & 1      & 3     \\
$C\delta_i$ = 1  & 20        & 4      & 16    \\
$C\delta_i$ = 2  & 31       & 7      & 24     \\
$C\delta_i$ = 3  & 16       & 2      & 14    \\
$C\delta_i$ = 4  & 4        & 1      & 3    \\

\bottomrule
\end{tabular}%
\caption{Average inter-annotator disagreements $\delta$ (Eqn~(\ref{eqn:delta})) the frequency of $C\delta_i$ (Eqn~(\ref{eqn:delta_i})) obtained from non-MTurk annotators. Higher scores mean more disagreement.}
\label{tab:bl_nat_disagree}
\end{table}

\featpar{Non-MTurk annotators.} In light of \citet{karpinska-etal-2021-perils} on poor performance of Turkers evaluating generated text (albeit open-ended), as a check on our Turker annotations we also recruited three colleagues who are native speakers of English, and not part of this work and unaware of the purpose of the study, to annotate a sample of 75 sentences ($5 \times 15$ sets).

Table~\ref{tab:bl_nat_disagree} shows that the agreement levels for this smaller sample mirrors Table~\ref{tab:annot_disagree}, except that the adversarial examples are all grouped together.  The overall $\delta$ of 0.65 is very similar to the MTurkers' 0.61 (and in fact, even representing slightly greater disagreement than the MTurker results), giving support to the results of Table~\ref{tab:annot_disagree} being representative of the task.
Also as with the MTurkers, there was better agreement (here only slightly) on original texts than on adversarial ones.

\featpar{Links to detection of adversarial examples.}
As another reliability check, we relate our data to the prior work of \citet{dyrmishi-etal-2023-humans}, where their suspiciousness regarding whether a text is human-generated or computer-altered is measured by a binary question.
An obvious possible transformation of our data to binary is to consider scores of 1 and 2 as non-suspicious or human-like and scores of 4 and 5 as suspicious or computer-generated; this reflects the textual instructions given to our annotators, and we refer to this as the symmetry-split. An alternative transformation, the 1-vs-other split, considers only the score of 1 as non-suspicious and all other scores as indicative of some degree of suspicion. Table~\ref{tab:suspicion_level} displays results from both types of splits. In the symmetry-split scenario, we would interpret this as saying that annotators only recognise 27\% of all adversarial texts as being computer-altered, which is ridiculously low. However, in the 1-vs-other scenario, the percentage of texts raising suspicion increases to 69\%, aligning more closely with the findings of \citet{dyrmishi-etal-2023-humans} (60.3\%). 

In terms of the rankings of adversarial methods by suspiciousness, we see a small difference in this 1-vs-other binarization versus the mean Likert score of Table~\ref{tab:sus_counts}.  Under the mean Likert scores, BAE is the worst, followed by \Pruthi (although they are not statistically significantly different), whereas this is reversed in 1-vs-other.  This is due to BAE having the largest number of 5 scores (most obviously computer-altered) in Likert counts by some margin, but it also does slightly better than \Pruthi on 1 score.  This more extreme variability of performance by BAE is to some extent concealed by the binarization.  \Alzantot and \Textfooler are closer to each other, and closer to humans than the other two, in both rankings. 

Our rankings have some differences from \citet{dyrmishi-etal-2023-humans}. 
 There, as here, \Alzantot was observed to generate the most human-like sentences --- only 46.5\% were suspected to be computer-altered there --- but the ordering of \Bae (50.6\%) and \Textfooler (61.5\%) are the reverse of ours.  We are not sure what factors might lead to this.
 \\

\begin{table}
\centering
\resizebox{0.7\columnwidth}{!}{%
\begin{tabular}{@{}lcccc@{}}
\toprule
           & \multicolumn{2}{c}{Symmetry split} & \multicolumn{2}{c}{1-vs-other split} \\\cmidrule{2-5} 
 &
  \begin{tabular}[c]{@{}c@{}}Human\\ (1,2)\end{tabular} &
  \begin{tabular}[c]{@{}c@{}}Computer \\ (4,5)\end{tabular} &
  \begin{tabular}[c]{@{}c@{}}Human \\ (1)\end{tabular} &
  \begin{tabular}[c]{@{}c@{}}Computer\\(2-5)\end{tabular} \\ \midrule
Overall    & 0.67            & 0.25           & 0.31             & 0.69            \\
\midrule
Real       & 0.76            & 0.17           & 0.39             & 0.61            \\
Adv        & 0.64            & 0.27           & 0.29             & 0.71            \\
\midrule
\Pruthi     & 0.60            & 0.30            & 0.24             & 0.76            \\
\Alzantot   & 0.71            & 0.21           & 0.34             & 0.66            \\
\Textfooler & 0.69            & 0.21           & 0.35             & 0.65            \\
\Bae        & 0.55            & 0.35           & 0.25             & 0.75            \\ \bottomrule
\end{tabular}%
}
\caption{Proportions of suspiciousness scores split in symmetry (scores of 1,2 human-authored and 4,5 computer-altered) and 1-vs-other (score 1 human-authored and 2-5 computer-altered). (Note that proportions under symmetry split do not add up to 1 as the score of 3 is excluded.)}
\label{tab:suspicion_level}
\vspace{-.8 cm}
\end{table}

\featpar{Impact of the survey instructions: prevalence.} 
A known effect in psychology is that humans can make judgements about items based on their belief about prevalence \citep[on the `blue dot effect']{levari-etal:2018:Science}. 
In our primary data-gathering setup, no information about prevalence is provided; this is reflective of our scenarios of interest (as, for example, when a person sees a single review with typos and speculates about whether it was authentically written by a human or modified).  However, we may also wonder how scores would be affected if human judges were provided with information about prevalence, and if this would change overall observed patterns.
In our setup, where annotators are not provided with information about prevalence, they may expect an even split between human and computer-altered sentences (which is also the usual setup in the work described above).
Such an assumption about prevalence might incline annotators towards identifying more adversarial sentences as real. To investigate whether this is the case, and see whether providing information about prevalence changes the patterns we have observed, we conducted a smaller survey with 750 sentences with the same distribution of original and adversarial sentences, consisting of 510 single annotations across 102 x 5 sets of sentences, and 3 annotations on 16 x 5 sets of sentences, explicitly instructing the participants that there were more computer-altered sentences than real ones. The results, presented in Table~\ref{tab:sus_counts_v2}, indicate that this change in instruction does not significantly alter the outcomes compared to those shown in Table~\ref{tab:sus_counts}. The annotators seem slightly more suspicious overall --- the mean score across all texts is 2.60 versus 2.35 under the main setting --- but the relative rankings are all the same, with original sentences scoring lowest (most human-like), \Pruthi and \Bae the most suspicious, and the other two methods in the middle.  Other patterns from the original setting likewise show themselves (e.g. scores of 3 and 5 chosen infrequently).  As well as indicating that instructions regarding prevalence do not matter particularly for assigning suspiciousness scores --- although they might for human detection of computer-altered text --- this second albeit smaller dataset supports the conclusions of the original.%
\footnote{For completeness, we also looked to see if telling annotators that there were fewer adversarial texts than original human ones affected the scores. This differs from our main prevalence experiment in that we are there, on data containing original to adversarial sentences in the ratio 1:4, contrasting the cases of no information about prevalence with true information about prevalence.  Here, using data with the same distribution means that we give them \textit{incorrect} information about prevalence. Consistently with the earlier results, annotators do take prevalence information into account when it is available (even if false): they are slightly less suspicious overall as measured by mean score, but with other patterns --- such as original sentences' mean scores being lower than all adversarial sentence mean scores --- remaining the same.}
\\

\begin{table}
\centering
\resizebox{0.7\columnwidth}{!}{%
\begin{tabular}{@{}lrrrrrc@{}}
\toprule
Sus. score  & 1    & 2    & 3   & 4   & 5   & Mean score \\
\midrule\midrule
Total     & 143 & 296 & 84 & 176 & 51 & 2.60     \\\midrule
Original    & 38  & 60 & 14  & 30 & 9  & 2.42 \\
Adversarial & 105  & 236 & 70 & 146 & 42 & 2.64     \\\midrule
\Pruthi      & 25  & 65  & 22  & 40 & 13  & 2.70     \\
\Alzantot    & 29  & 63  & 19  & 33 & 6  & 2.49     \\
\Textfooler  & 30  & 56  & 19  & 34 & 10  & 2.58     \\
\Bae         & 21  & 52  & 10  & 39 & 13  & 2.79     \\ \bottomrule
\end{tabular}%
}
\caption{Frequency of each Likert-scale suspicion score $[1 \ldots 5]$ for original human and adversarial sentences provided to the human annotators, along with mean suspicion scores, under the prevalence-informed setting. ``Adversarial'' represents the total of all adversarial methods.}

\label{tab:sus_counts_v2}
\end{table}

\featpar{Correlation with text evaluation metrics.}  
As we observed in Section~\ref{sec:RelatedWork},
adversarial example generation methods generally implement some constraints on various properties to improve example quality which include semantic and other similarity to the original, measured by a range of metrics.  However, these methods do not consider the non-suspicion as a novel characteristic to be measured. We therefore examine the extent to which our non-suspicion scores are in fact distinct from the kinds of metrics used to measure overlap with the original sentence.
Table~\ref{tab:quality_suspicion} presents the average scores for BLEU \cite{papineni-etal-2002-bleu}, METEOR \cite{banerjee-lavie-2005-meteor}, ROUGE \cite{lin-2004-rouge}, BERTScore \cite{zhang2020bertscore} MeaningBERT \cite{david2023meaningbert} and BLEURT \hbox{\cite{sellam-etal-2020-bleurt}} of the adversarial sentences, alongside their correlation with human suspicion ratings. We see that the non-suspicion scores have almost no relationship to other metrics, indicating it is in fact a different dimension of adversarial example quality.

\begin{table}
\centering
\resizebox{0.9\columnwidth}{!}{%
\begin{tabular}{@{}lcccccc@{}}
\toprule
Metrics         & \Pruthi & \Alzantot & \Textfooler & \Bae   & Overall  & $r$  \\ \midrule
BLEU            & 0.869  & 0.752    & 0.827      & 0.838 & 0.856 & -0.057 \\
METEOR          & 0.947  & 0.908    & 0.935      & 0.935 & 0.945 & -0.064 \\
ROUGE1          & 0.941  & 0.882    & 0.922      & 0.928 & 0.934 & -0.064 \\
ROUGE2          & 0.893  & 0.794    & 0.859      & 0.869 & 0.882 & -0.060 \\
ROUGE3          & 0.844  & 0.713    & 0.798      & 0.810 & 0.831 & -0.058 \\
ROUGEL          & 0.941  & 0.882    & 0.922      & 0.928 & 0.934 & -0.064 \\
BERTScore\textsubscript{F1}   & 0.985  & 0.985    & 0.986      & 0.984 & 0.985 & -0.094 \\
BERTScore\textsubscript{Prec} & 0.981  & 0.984    & 0.985      & 0.983 & 0.983 & -0.106 \\
BERTScore\textsubscript{Rec}  & 0.989  & 0.986    & 0.988      & 0.985 & 0.987 & -0.074 \\
MeaningBERT     & 0.877  & 0.878    & 0.877      & 0.875 & 0.877 & 0.023  \\
BLEURT          & 0.126  & 0.489    & 0.448      & 0.159 & 0.428 & -0.138 \\ \bottomrule
\end{tabular}%
}
\caption{Mean metric score between pairs of original and adversarial sentences generated by each method. ``Overall'' represents scores across all methods.  $r$ is the Pearson correlation coefficient with respect to human suspicion scores.}
\label{tab:quality_suspicion}
\end{table}

\section{Predicting the Level of Suspiciousness}
\label{sec:sus_reg}

In this section, we look to build a predictor of the scores described in Section~\ref{sec:sus_annotation}.
In doing this, we look at other regression tasks based on subjective opinions such as the evaluation of abstractness through human imperceptibility in SemEval 2021 \citep{zheng-etal-2021-semeval} and the assessment of semantic textual similarity in SemEval 2017 \hbox{\citep{cer-etal-2017-semeval}}. The SemEval 2021 task involves rating words based on their abstractness, while the SemEval 2017 task focuses on evaluating multilingual and cross-lingual sentences' semantic similarity on a 0-5 Likert scale. We thence build several regressors that are trained using scores from human annotations to predict human suspicion scores on adversarial texts.

\subsection{Experimental Setup}
\label{sec:reg_exp_setup}

\featpar{Likert score dataset.} 
Our primary dataset for our prediction models is the dataset described in Section~\ref{sec:sus_annotation}, consisting of 3015 total annotated sentences from the \MR dataset.  From this, we take the 315 sentences with 3 annotations each as the test set and use the median as the assigned Likert score for the item.  Of the remaining 2707 (single annotation) sentences, we split them 90-10\% as train and dev.
\\

\featpar{Morris dataset.}
\citet{morris-etal-2020-reevaluating}, as noted above, collected binary suspicion judgements on a small set of 100 unique sentences from the \MR dataset and their \Textfooler generated version (so 200 texts in total).  As each of these was judged by 10 annotators, we can derive a scalar score from this by taking the average of the annotations, treating human as 0 and computer-altered as 1; the score is thus between 0 (all annotators think computer-altered) and 1 (all annotators think human), with a score of 0.5 indicating most uncertainty. We use 144, 20 and 36 sentences as the train, dev and test set.  This dataset, due to its small size, is very much secondary and only used as a check on our Likert dataset. Table \ref{tab:ex_morris}, with its illustrations of a few sentences from the Morris dataset, includes our derived suspicion score in the rightmost column.%
\footnote{It is not possible to create a similar dataset from \citet{dyrmishi-etal-2023-humans}, as almost all of their sentences have only a single binary judgement.}
\\

\featpar{Evaluation metrics}
Like \citet{cer-etal-2017-semeval}, we consider Pearson's correlation to be our primary evaluation metric, but we also report Spearman's rank correlation and root-mean-squared-error (RMSE) scores.%
\footnote{Multiple metric perspectives can be useful. Consider Pearson's $r$ and Spearman's $\rho$ \citep{van-den-heuvel-zhan:2022}: the former is considered a measure of linear association, while the latter is considered a measure of monotonic association.  While they give similar scores on normally distributed data, on non-linear but monotonic data $\rho$ scores are systematically higher.  In addition, $\rho$ is less sensitive to outliers; consequently, if there are major outliers, $\rho$ will differ from $r$.}

\subsection{Regressor models}

The core of our regression models is a language model over the texts.  However, following the best regression models in Semantic Textual Similarity SemEval task \citep{cer-etal-2017-semeval}, we use additional numerical features.  We then combine these together in two ways:
1) \textbf{\Combined}, combining the prediction of two regressors separately trained on texts and numerical features  by averaging their predictions; and 2) \textbf{\Multi}, training multimodal regressors on both text and numerical features using multimodal Transformers 
as proposed by \citet{gu-budhkar-2021-package}.

\paragraph{\textbf{Language model component}}
For these, we fine-tune BERT\textsubscript{BASE}, RoBERTa, XLNET, DistilBERT, and ALBERT. 
In addition, we fine-tune BERT and RoBERTa models using the SMART approach of \citet{jiang-etal-2020-smart}.
SMART incorporates smoothness-inducing regularization and Bregman proximal point optimization techniques, preventing overly aggressive updates and ensuring the model remains within a small neighborhood of its previous iteration at each step. SMART achieves the highest performance on the STS semantic similarity benchmark \citep{cer-etal-2017-semeval}\footnote{\url{https://paperswithcode.com/sota/semantic-textual-similarity-on-sts-benchmark}} and is top-performing in the Adversarial GLUE multi-tasking benchmark \citep{wang2021advglue}
for evaluating language model robustness. We refer to these models as \SMARTB and \SMARTR.

All the regressors are trained from 1 to 10 epochs with AdamW optimizer 0.01 weight decay. We choose the regressors having the maximum Pearson's correlation on the devset. For the Morris dataset, just consider BERT, RoBERTa, \SMARTB and \SMARTR regressors for comparison.

\subsection{Numerical Features}
\label{sec:reg_features} 

We use the features below and train several models on them for use in the \textbf{\Combined} setup: LinearRegressor, HuberRegressor, RandomForestRegressor, GradientBoostingRegressor, XGBoostRegressor, VotingRegressor and StackingRegressor. In these models, all numerical features are normalised using quantile normal transformation.

\begin{itemize}
\item 
\featpar{Rate of perturbations.} This feature is represented by the percentage of words modified by the adversarial attack on a sentence. \textbf{(1 feature)}

\item 
\featpar{Word embeddings.} We use GoogleNews word2vec \citep{NIPS2013_9aa42b31} vectors, with 300 dimensions. We represent each sentence by concatenating the average pooled vector representations of its words. Additionally, each word vector is weighted by its Inverse Document Frequency (IDF) value.  We take each element of the vector as a feature. \textbf{(300 features)}

\item 
\featpar{Adversarial text detectors.} Automated adversarial text detection has shown notable success \citep{wang2022detecting,mosca-etal-2022-suspicious,bao2021defending,nguyen-son-etal-2023-votetrans,tonni-dras-2023-IJCNLP-learned}. 
Several of these methods use learned representations to identify properties that distinguish human-authored texts from adversarial ones.
Incorporating features from these detectors may enhance our suspicion regressor's capabilities. From \citet{tonni-dras-2023-IJCNLP-learned}, we utilise the influence function (IF) score, and \dknn based useful and harmful ranks and distances, as used in the effective \NNIF detector\footnote{\NNIF investigates a combined viewpoint of identifying the neighborhood subspace of the adversarial sentences based on influence function \citep{koh2017understanding} and \dknn \citep{papernot2018deep}, following \citet{CVPRPaper}. Interested readers can refer to \citet{tonni-dras-2023-IJCNLP-learned}.} \textbf{(5 features)}. Additionally, from \citet{liu-etal-2022-detecting:RepL4NLP}, we calculate the local intrinsic dimensionality (LID) scores for each layer of BERT as well as the average LID. \textbf{(13 features)}

\item 
\featpar{Grammaticality scores.} Grammatical accuracy of these perturbations may correlate with human suspicions. We use both the GramFormer\footnote{\url{https://github.com/PrithivirajDamodaran/Gramformer/tree/main}} and LanguageTool (adapted from \Textattack) packages. \textbf{(2 features)}  

\item 
\featpar{GPT-4 produced scores.} \citet{chiang-lee-2023-large} use LLMs to produce human-like scoring on texts and show that ChatGPT in particular chooses human written texts over adversarial texts in evaluating fluency
and meaning preservation. 
Thus, besides collecting human annotations, we also design a GPT LLM (GPT-4 pointing to GPT-4-0613) using the OpenAI API, to predict scores of suspicion level on a scale of 1--5, and give a reason. The instruction contains three example sentences along with explanations of their scoring (App. table \ref{tab:gpt_prompt}). Appendix Table~\ref{tab:gpt_response} presents a few examples of suspicion scoring and reasoning by ChatGPT. \textbf{(1 feature)}

\end{itemize}

Overall, this gives 322 numerical features.

\begin{table}[t]
\centering
\resizebox{\columnwidth}{!}{
\begin{tabular}{@{}llccc@{}}
\toprule
Regressor type& LM         & Pearson's r $\uparrow$ & Spearman's $\rho$ $\uparrow$ & RMSE $\downarrow$ \\ \midrule

\Text &            BERT          & 0.396         & 0.323        & 1.311 \\
 &           RoBERTa       & 0.336         & 0.310        & 1.220 \\
 &           AlBERT        & 0.119         & 0.123        & 2.242 \\
 &           DistilBERT    & 0.192         & 0.198        & 2.613 \\
 &           XLNet         & 0.074         & 0.071        & 2.580 \\
 &           \SMARTB    & 0.377        & 0.322         & 1.342 \\
 &           \SMARTR & 0.362        & 0.323         & 1.259 \\\midrule

\Numeric &           VotingReg.& 0.245         & 0.249       & 1.259 \\
 &           StackingReg.& 0.120        & 0.129        & 1.331 \\
 &           GradientBoosting& 0.195       & 0.198        & 0.234 \\
 &           RandomForest& 0.456       & 0.408        & 1.235 \\
 &           XGBoost& 0.112       & 0.133        & 1.350 \\ 
 &           LinearReg.& 0.214       & 0.218        & 1.322 \\
 &           HuberReg.& 0.183       & 0.189        & 1.402 \\ \midrule

\Combined &            BERT+RandomForest  &    \textbf{0.515}         & \textbf{0.585}            &  1.125    \\\midrule

\Multi &           BERT          & 0.301       & 0.303        & 1.219 \\
 &           RoBERTa       & 0.272       & 0.269        & 1.244 \\
 &           AlBERT        & 0.141       & 0.184        & 1.368 \\
 &           DistilBERT    & 0.061       & 0.081        & 1.969 \\
 &           XLNet         & 0.216       & 0.210        & 2.016 \\
 &           \SMARTB    & 0.274       & 0.274        & 1.264 \\
 &           \SMARTR & 0.173       & 0.196        & 1.318 \\
            \bottomrule
\end{tabular}
}
\caption{Pearson's and Spearman's correlation coefficients and the RMSE values between the various regressor scores and the human suspicion scores. ($\uparrow$ indicates higher and $\downarrow$ indicates lower is better)}
\label{tab:single_sent_reg}
\end{table}

\subsection{Results and Discussion}
\label{sec:reg_result_analysis}

\featpar{Overall.} Table~\ref{tab:single_sent_reg} presents the Pearson and Spearman correlation coefficients, along with the RMSE scores.  Overall, there is a moderate correlation (Pearson's $r = 0.515$ for the best model), demonstrating that this is a challenging task; we expect this is a reflection of the level of agreement among annotators discussed in Section~\ref{sec:sus_annotation}. The Pearson and Spearman coefficients are close, so we do not see any evidence of major outliers, etc.
\\

\featpar{Ensembling separately trained text and numerical models perform better (Table~\ref{tab:single_sent_reg}).}
From Table \ref{tab:single_sent_reg}, using only the texts, we achieve $r$ = 0.396 with a \textsc{BERT} model and a closely related outcome of 0.377 with \SMARTB. Among the ML regressors focusing solely on numerical features, \textsc{RandomForest} achieves the highest Pearson’s $r$ at 0.456. We enhance predictive accuracy by ensembling the predictions from these two models in a \Combined setting following \citet{tian-etal-2017-ecnu}.
The best correlation with $r = 0.515$ is achieved by combining predictions from the best \Combined model. 
Using the multimodal setting incorporating both texts and numerical features, in contrast, the highest Pearson’s $r$ achieved is 0.301.
\\

\featpar{Not all numeric features perform equally (Table~\ref{tab:feat_subset}).} Additional experiments on the subsets of the numerical features show that combining some of the features with the texts gives better results than the others.
We experiment with training the regressors using the \NNIF features, grammaticality scores by GramFormer, LanguageTool and GPT-4 generated scores paired with the \Text and the results are illustrated in Table~\ref{tab:feat_subset}. There is a significant level of difference in the regressor outcome and adding the GPT scores (\Text+GPT and \Text+\NNIF+GPT) with the text improves the result a lot.
\\

\begin{table}
\centering
\resizebox{.8\columnwidth}{!}{
\begin{tabular}{@{}lcccc@{}}
\toprule
 \Text+&
  BERT & RoBERTa &
  \SMARTB &
  \SMARTR \\ \midrule
\NNIF          & 0.272 & 0.282 & 0.287 & \textbf{0.323} \\
\NNIF+Gram.      & 0.289 & 0.288 & 0.289 & 0.322 \\
\NNIF+Lang.    & 0.272 & 0.272 & 0.244 & 0.307 \\
GPT            & 0.351 & \textbf{0.407} & \textbf{0.382} & 0.221 \\
GPT+Gram.        & 0.280 & 0.306 & 0.280 & 0.280 \\
\NNIF+GPT     & \textbf{0.386} & 0.218 & 0.335 & 0.233 \\ 
\NNIF+GPT+Gram.& 0.320 & 0.242 & 0.316 & 0.264 \\
\bottomrule
\end{tabular}}
\caption{Pearson's $r$ obtained by the \Multi regressors using text and different sets of features. Higher is better.}
\label{tab:feat_subset}
\end{table}

\begin{figure}
\centering
\normalsize
\includegraphics[width=0.7\columnwidth]{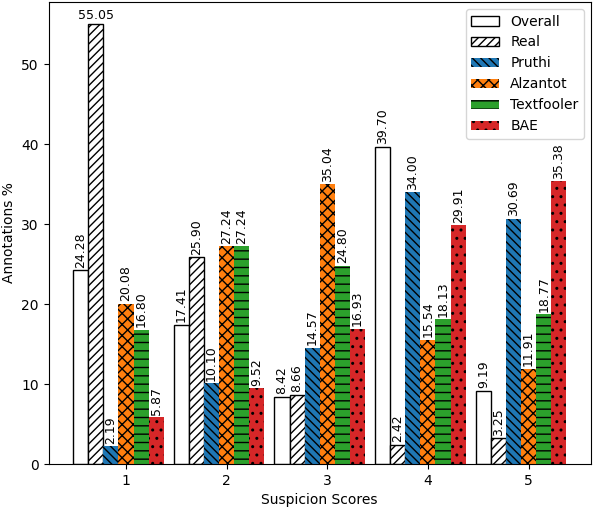}
\caption{Distribution (\%) of the ChatGPT (GPT-4) annotations grouped by the Likert scale levels.
}
\label{fig:annot_dist_gpt}
\end{figure}

\featpar{GPT-4 as a suspicion scoring baseline (Figure \ref{fig:annot_dist_gpt})}. Although GPT-generated scores work well as a training feature, the distribution of the suspicion levels (ranging from 1 to 5) determined by ChatGPT in Figure \ref{fig:annot_dist_gpt} suggests that it works better as an adversarial detector than in predicting human suspicions scores as such.
\\

\featpar{Regressors trained on subsets of the data (Table~\ref{tab:reg_subset})}.
We also examined how well regressors performed that were trained on the data subset by type of text (original text and each adversarial method). The Pearson's correlation coefficients obtained from regressors trained just on BERT textual representations are presented in Table~\ref{tab:reg_subset}. 
Interestingly, the regressor trained on original text was best not only overall and specifically against other original text but also the best against \Pruthi (by a large margin) and \Textfooler, even better than the regressors trained on that data.  This suggests that better predictors of adversarial example suspiciousness could be built counter-intuitively by gathering more data only on original human text.
\\

\begin{table}
\centering
\resizebox{\columnwidth}{!}{
\begin{tabular}{@{}lcccccc@{}}
\toprule
      & \multicolumn{6}{c}{Test subset} \\ \cmidrule{2-7}
     Train subset & Overall & \textsc{Original}& \Pruthi & \Alzantot & \Textfooler & \Bae   \\\midrule
     All& 0.52 & 0.50 & 0.49 & 0.22 & 0.37 & 0.15\\\midrule
    \textsc{Original}       & \textbf{0.30}   & \textbf{0.45}    & 0.43 & 0.15   & 0.29     & 0.08 \\
    \Pruthi        & 0.22   & 0.08    & 0.05  & 0.20    & 0.24      & \textbf{0.25} \\
    \Alzantot      & 0.16   & 0.08   & 0.07  & 0.16   & 0.02      & 0.01\\
    \Textfooler    & 0.15   & 0.07 & 0.11  & \textbf{0.27} & 0.04     & 0.13 \\
    \Bae           &  0.17   & 0.31    & 0.30  & 0.20    & \textbf{0.32}      & 0.07 \\ \bottomrule 
\end{tabular}}
\caption{Pearson's $r$ obtained by the regressors trained only on each subset of the train data and tested on all test data and the subsets. For comparison, the values when all data were used for training are also included. Higher is better.}
\label{tab:reg_subset}
\vspace{-0.5 cm}
\end{table}

\featpar{Results on Morris dataset.} We achieved similar although slightly better correlations with human judgements using regressors built on the Morris dataset utilizing only textual data. Specifically, we achieved a Pearson's correlation coefficient $r$ of 0.512, 0.506, 0.495 and 0.594 using BERT, RoBERTa, \SMARTB and \SMARTR, respectively.  While not too much can be made of the results on such a small dataset, it is possible that collecting a large number of binary judgements and deriving scores from them would produce more systematically predictable scores.  This would, however, be more expensive than collecting Likert scores.
\\

\section{Generating Low-suspicion Adversarial Texts}

A possible application of the regressors presented in Section~\ref{sec:sus_reg} is to use them to create adversarial texts that are less obvious to humans.  Such an application would be another way of assessing the effectiveness of the regressors and the dataset used to create them.
In this section, to illustrate this kind of application, we build an adversarial example generator that incorporates a non-suspicion constraint, with the aim that the examples generated will look less suspicious to humans.  We compare the original adversarial examples against their suspicion-constrained (\Susconstrained) variants in a human evaluation and show that the suspicion-constrained ones are perceived as better.

\subsection{Experimental Setup}

\featpar{Generator.}
We selected the \Textfooler method in the \Textattack framework as our base generator, as it has the highest number of Likert score 1 ratings (i.e., most human-like) in our human evaluations (Table~\ref{tab:sus_counts}), as well as being most effective in fooling a classifier (Table~\ref{tab:bert_clean_adv}). 

We then produced a variant of this incorporating a suspicion constraint that requires the algorithm to generate adversarial sentences that maintain suspicion levels below a specified threshold; this variant thus has all of the constraints in \Textfooler plus this one additional one. Suspicion scores for each candidate sentence are determined using our best BERT-based \Combined regressor (Table~\ref{tab:single_sent_reg}).

There are many possible ways to set a threshold for the constraint, for example as a function of the difference between the original text's regressor score and that of its \Textfooler adversarial counterpart.  We choose the simplest kind of threshold, a fixed value.
\\

\featpar{Selected texts.}
We considered how many texts from the test set are available under two different thresholds: $\tau=2.5$ and $\tau=3.0$.  We used $\tau = 2.5$ as our starting point in order to capture particularly bad \Textfooler texts; this is above the mean Likert scores for original and \Textfooler texts from Table~\ref{tab:sus_counts}.
We then examined which texts could have their predicted suspicion score reduced.  For $\tau = 2.5$, there are 61 \Textfooler texts with regressor scores $>2.5$, and 42 (80.3\%) of those can have their suspicion score reduced, whereas a threshold of $\tau=3.0$ results in only 27 \Textfooler texts of which 20 (74.1\%) can be reduced.  This latter is too small, so we chose $\tau=2.5$ for the experiment.  The mean reduction in predicted suspicion score for this threshold was $0.63$, with a standard deviation of $0.52$.

Since we are using just a simple single fixed threshold for the maximum suspicion score, adding the constraint can drive the suspiciousness even below the suspiciousness score of the original text; this might flag an implausible score, as it is unlikely that \Textfooler perturbation instances would make a sentence more human-like than the original human-authored one.  On the other hand, we do not want to necessarily remove all whose scores are below the original, as this might result in too few sentences to test.  Heuristically, we consider only those \Susconstrained variants whose scores fall below the corresponding original text's score by 0.2, which is half the gap between original and \Textfooler text mean scores.
(A more sophisticated constraint system could instead include, for example, a lower bound on non-suspicion that is a function of the original text.)  This gave 23 texts.
\\

\featpar{MTurk setup.}
Using the 23 selected texts, we conducted two subsequent surveys on Amazon Mechanical Turk to compare the quality of \Textfooler and \Susconstrained texts. For the first, we structured the survey to present text pairs, each consisting of a Textfooler sentence and its suspicion-constrained counterpart, and asked the annotators to determine which sentence of the pair appeared to be written by a human: that is, it was a forced-choice preference testing setup. We gathered 5 judgements on each sentence pair and took the overall judgement on the majority vote. For the second survey, to see the effect on Likert score ratings for comparability with Section~\ref{sec:sus_annotation}, we gathered the ratings in the same manner as in Section~\ref{sec:annot_setup} (i.e., Turkers were presented with individual texts in random order and asked to rate on the provided scale, etc).

\subsection{Analysis}

We consider first the forced-choice data.
For each of the 23 pairs of texts, in 18 cases the \Susconstrained text was preferred and in 5 cases the \Textfooler one.  Exact binomial tests are commonly used across many fields to test for significance in this kind of preference testing setup \citep[for example]{omahoney:1986}; we use the usual $\alpha = 0.05$.  The two-sided test result in this case is $p = 0.0106$, indicating that in our small sample, the \Susconstrained texts are significantly better.

For the Likert scale data, the \Textfooler texts had a mean Likert score of 2.61\footnote{Recall that the \Textfooler texts were selected to be ones that were ``particularly bad'', i.e., had high predicted score from our regressor, and so we would expect this to be higher than the mean for all \Textfooler texts (2.24) from Table~\ref{tab:sus_counts}.} versus 2.26 for \Susconstrained texts, a difference in mean score of 0.35.  For context, this improves our particularly bad \Textfooler texts to score approximately the same as the overall \Textfooler texts (mean Likert score 2.24) from Table~\ref{tab:sus_counts}; for an alternative perspective, it is approximately the same difference as between human-authored sentences (mean 2.05) and adversarial sentences overall (mean 2.42) in Table~\ref{tab:sus_counts}.

\begin{table*}
\centering
\resizebox{\textwidth}{!}{%
\begin{tabular}{@{}lp{13cm}cc@{}}
\toprule
 &\multicolumn{1}{c}{Scenario \#1: \Susconstrained $>$ \Textfooler: 5 $-$ 0 votes} & Human & Regr.\\\midrule
Orig & the skirmishes for power waged among victims and predators settle into an undistinguished rhythm of artificial suspense & 1 & 1.76\\
TF& the \diffcolour{intermingled}  for power waged among victims and predators settle into an undistinguished rhythm of \diffcolour{synthetic awaited} & 4 & 3.36\\
SC& the \diffcolour{struggles} for power \diffcolour{realized with} victims and predators settle into an undistinguished rhythm of \diffcolour{synthetic} suspense& & 1.83 \\
\midrule
&\multicolumn{1}{c}{Scenario \#2: \Susconstrained $>$ \Textfooler  : 3 $-$ 2 votes} &&\\\midrule
Orig & oliveira  seems to  pursue  silent  film  representation with every mournful  composition.
& 2 & 2.13 \\
TF & oliveira seems to  \diffcolour{sues}  silent  \diffcolour{image}  representation with every mournful composition.
& 4 & 2.76 \\
SC& oliveira seems to  \diffcolour{undertaken}  silent  \diffcolour{image}  representation with every mournful composition.
& -& 2.46 \\
\midrule
&\multicolumn{1}{c}{Scenario \#3: \Susconstrained $<$ \Textfooler: 2 $-$ 3 votes} &&\\\midrule
Orig &it's absolutely amazing how first-time director kevin donovan managed to find something new to add to the canon of chan. make chan's action sequences  boring.
 & 2 & 2.51 \\
TF& it's absolutely amazing how first-time director kevin donovan managed to find something new to add to the canon of chan. make chan's action sequences \diffcolour{uninspiring}. 
 & 2 & 3.17 \\
SC&  it's absolutely amazing how first-time director kevin donovan managed to find something new to add to the canon of chan. make chan's action sequences \diffcolour{tedium} & - & 2.38 \\ \bottomrule
\end{tabular}%
}
\caption{Three scenarios consisting of an original sentence, a \Textfooler (TF) variant, and a \Susconstrained (SC) variant.  For each sentence, the human Likert score (``Human'') --- not available for SC, by definition --- 
and the regressor-predicted score (``Regr.'') are provided. In Scenarios \#1 and \#2, \Susconstrained was chosen over \Textfooler, based on votes. In Scenario \#3, \Textfooler was chosen to be a better variant.}

\label{tab:sent_cases}
\end{table*}

Table~\ref{tab:sent_cases} gives some examples: it notes the vote among the 5 annotators for the \Susconstrained and plain \Textfooler versions, the human Likert scores (for original and \Textfooler) and predicted scores (for all three texts). In the first case, all annotators preferred the suspicion-constrained version, for which the suspiciousness score predicted by the regressor is much better than the \Textfooler one; the strength of the vote (5-0) is not surprising, given how bad the \Textfooler variant is compared to the original, both in human and predicted scores (1 vs 4, 1.76 vs 3.36), and the closeness of the \Susconstrained's predicted score to the original (1.76 vs 1.83).  
In the second case, the vote was marginal in favour of the \Susconstrained version.  Here, the predicted suspiciousness score for the suspicion-constrained version is about halfway between the original and the \Textfooler versions, a smaller relative improvement than in the first case  (2.76 vs 2.46).  
In the third case, it is the \Textfooler version that is preferred by a vote of 3 to 2.  Here, the \Susconstrained version actually has a predicted score that is lower than the original  (2.51 vs 2.38), which is an indication that the score is perhaps not likely to be accurate in this case; a more sophisticated constraint setup might have ruled this one out.

\section{Limitations}

A key limitation in this work is that we considered only character-level and word-level attacks, which are available via the \Textattack implementation; these types of attacks are well-established and have been around for a number of years. 
Paraphrase-level adversarial attacks represent an intriguing future area of research, especially with the advent of Large Language Models (LLMs). However, there are as yet only a very limited number of them, and none of them are suitable for our needs due to various limitations, as discussed in Section~\ref{sec:relwork-adv-methods}. Further, if they were to have non-suspicion constraints applied, they would either need to be couched within a framework like \Textattack, or some other means of application would have to be found.
Nevertheless, even though one might expect paraphrase attacks to produce more human-like and consequently less detectable texts, paraphrase work like \citet{lei-etal-2022-phrase} demonstrates in their human evaluation that their paraphrases were less grammatical than human-authored sentences, and performed similarly to the word-based attack CLARE. This suggests our notion of graded suspicion could potentially be relevant for paraphrastic attacks in the future as well.

In addition, like the work of \citet{morris-etal-2020-reevaluating} and \citet{dyrmishi-etal-2023-humans} that we use as our starting point, this work has only considered sentiment data, and only for English.  This is relatively easy to attack; on the other hand, there is perhaps more scope for variation in the quality of adversarial attacks for more challenging tasks, suggesting suspicion measures may be even more relevant on other datasets.  Further, we have only explored adversarial attacks within a classification paradigm, rather than generation \citep{zou2023universaltransferableadversarialattacks}, for example.

\section{Conclusions}

In this work, we investigated human judgements of the suspiciousness of adversarial texts, moving beyond binary classifications to capture varying levels of suspicion through a nuanced, Likert-scale approach; as part of this, we have made available a novel dataset of suspiciousness judgements across a range of adversarial methods. Additionally, we examined a range of regression-based models for predicting these judgements, to serve as a baseline for future research on automatically determining how suspicious humans will find adversarial text.  We further illustrate the utility of this scorer in imposing suspicion thresholds to enhance the generation of less suspicious adversarial texts. 

There are many avenues for future work in understanding and capturing human notions of what makes adversarial text suspicious.
As noted above in Limitations, extending adversarial methods to LLMs and tasks --- both classification tasks like Natural Language Inference used in the adversarial attack literature \citep{tonni-dras-2023-IJCNLP-learned}, or broader tasks like Named Entity Recognition \citep{srinivasan-vajjala-2023-multilingual} --- would be important.
Beyond this, other kinds of judgements, such as multiple binary judgements gathered on a large scale, could lead to improved models for prediction.  More recent LLM work, such as chain-of-thought prompting, could form an alternative approach to scoring, or at least to useful numerical features to use in the sorts of regression models presented in this paper.  
Finally, there is much work to be done to understand how these sorts of scores can be used as constraints to produce more human-like adversarial examples.

\clearpage
\appendix

\begin{figure}[ht]
    \centering
    \includegraphics[width=\textwidth]{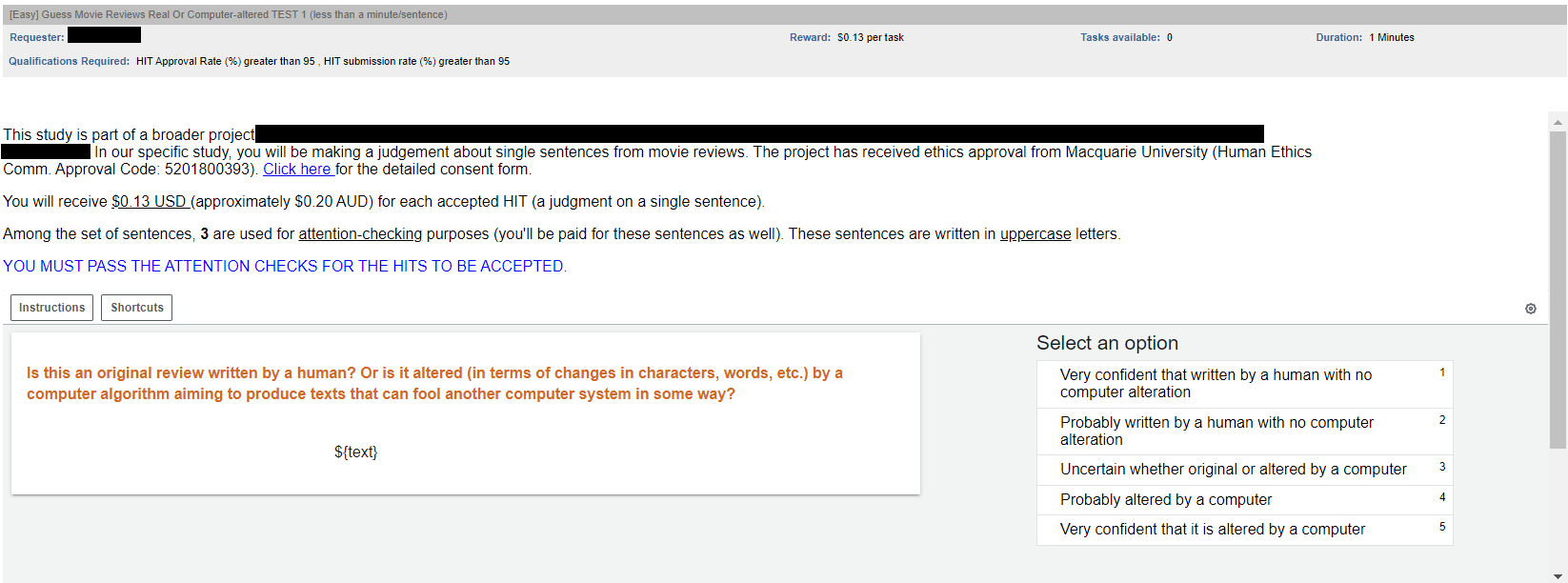}
    \caption{Amazon Mechanical Turk UI for collecting human judgements.}
    \label{fig:mturk_UI}
\end{figure}

\appendixsection{Human Evaluation Instruction}
The Amazon Mechanical Turk survey interface for human evaluation is illustrated in Figure \ref{fig:mturk_UI}. 

\appendixsection{ChatGPT Prompt}
The prompt used for the ChatGPT baseline is in Table \ref{tab:gpt_prompt}. A few suspicion-scoring responses on the original and adversarial texts are illustrated in Table \ref{tab:gpt_response}

\begin{table}[ht]
\centering
\resizebox{\textwidth}{!}{%
\begin{tabular}{@{}l@{}}
\toprule
"You are a highly intelligent AI trained to detect if a sentence might have been \\altered by a computer algorithm. "\\
"In the sentence, one or more words or characters might be altered by a computer\\thm to change another algorithm's prediction on it."\\
"However, sometimes character-level errors can be human errors as well." \\
"So when judging a sentence you need to see if the
sentence raises any suspicion \\looking at whether the words go with each other. On a scale of 1-5, where,"\\
"1 is "no computer alterations",\\
2 is "possibly no computer alteration", \\
3 is "uncertain whether altered or not",\\
4 is "possibly altered" and \\
5 is "certainly computer-altered", \\
you will generate a score for the suspiciousness of a sentence." \\

"For example, "it may . . . wok as a jaunt down memory 
lane for teens and\\ young adults" has a score of 4 as 
there is a character attack on the word "work"\\ and the 
letter "r" is deleted, but also can be a human error."\\ 

"Another example is "a cliched and surface cautionary 
tale about the hard-partying \\lives of teenagers ." having
a score of 5 as the word "shallow" is replaced by
another \\synonymous word "surface" and "surface cautionary"
doesn't make any \\sense in this context."\\

"Another example is "This is a great day to be alive." having a score of 1 as the characters \\and words are in place
in a meaningful order."\\
"Produce a numeric score of suspiciousness and reasoning in one line for a given sentence."\\
\bottomrule
\end{tabular}%
}
\caption{Used prompt for the OpenAI ChatGPT annotations.}
\label{tab:gpt_prompt}
\end{table}

\begin{table}[ht]
\centering
\resizebox{\textwidth}{!}{%
\begin{tabular}{@{}l@{}}
\toprule
\textbf{Original \#} a romantic comedy  yes  but one with characters who think and talk about \\their goals  and are working on hard decisions .\\
\\

\textbf{ChatGPT-response \#} Score: 1. The sentence appears to be perfectly logical, and all words\\ and characters are used appropriately in the context provided. There is no reason to suspect \\any alterations were made by an algorithm.\\\midrule

\textbf{Original \#} a film that will probably please people already fascinated by behan but leave everyone\\ else yawning with admiration .\\
\\
\textbf{ChatGPT-response \#} Score: 4. The phrase "yawning with admiration" seems suspicious because \\it combines two acts that are usually mutually exclusive. One does not generally yawn when \\they admire something. This could likely be a computer alteration, replacing a word with "yawning", \\but it could also be a less common choice of words from a human reviewer\\\midrule

\textbf{\Textfooler\#} a disturbing and frighteningly \diffcolour{plaintive} \underline{(orig:evocative)}  assembly of imagery and \\\diffcolour{melodious} \underline{(orig:hypnotic)} music composed by philip glass .\\
\\
\textbf{ChatGPT-response \#} Score: 1. The sentence is intact and correctly ordered. There isn't any \\out-of-place word, character or synonymous words replacing critical phrases that could raise \\suspicions of a computer-altered attack.\\\midrule

\textbf{\Textfooler\#}  the \diffcolour{intermingled} \underline{(orig:skirmishes)} for power waged among victims and predator \\settle into an undistinguished rhythm of \diffcolour{synthetic awaited} \underline{(orig:artificial suspense)} .\\
\\
\textbf{ChatGPT-response \#} I would give this sentence a score of 5. The sentence is full of seemingly \\unrelated words strung together in a nonsensical sequence, suggesting it was certainly altered \\by a computer algorithm. Especially the part "synthetic awaited" doesn't make sense in this context, \\indicating a high probability of computer alteration.\\\bottomrule

\end{tabular}%
}
\caption{LLM (GPT-4) generated suspiciousness scores on two original sentences and two \Textfooler sentences.}
\label{tab:gpt_response}
\end{table}

\begin{acknowledgments}

This project was undertaken with the assistance of resources and services from the National Computational Infrastructure (NCI), supported by the Australian Government. This project was also supported by the International Macquarie University Research Excellence Scholarship.

The human evaluation section of this study has received ethics approval from Macquarie University (Human Ethics Comm. Approval Code: 5201800393).
\end{acknowledgments}

\clearpage
\newpage

\setlength{\bibsep}{0pt}

\bibliography{anthology1,custom}

\end{document}